\documentclass[10pt, logo, onecolumn]{shengshu}

\usepackage[utf8]{inputenc}
\usepackage[T1]{fontenc}
\usepackage{charter}

\newcommand{\bffont}{\fontsize{8.8}{10}\selectfont}
\DeclareTextFontCommand{\textbf}{\bffont\bfseries\selectfont}

\usepackage{wrapfig}
\usepackage{xspace}
\usepackage{textgreek}
\usepackage{bm}
\usepackage{cleveref}
\usepackage{multirow}
\usepackage{subcaption}
\usepackage{orcidlink}
\usepackage{footmisc}
\usepackage{algorithm}
\usepackage{algpseudocode}
\usepackage{listings}
\usepackage{csquotes}
\usepackage{marvosym}
\usepackage{nicefrac}
\usepackage{multicol}
\usepackage{tabto}
\usepackage{adjustbox}
\usepackage{dblfloatfix}
\usepackage{verbatim}
\usepackage{mathtools}
\usepackage{array}
\usepackage{bbm}
\usepackage{makecell}
\usepackage{siunitx}
\usepackage{pdflscape}
\usepackage{arydshln}
\usepackage{hhline}
\usepackage{diagbox}
\usepackage[square,sort,comma,numbers]{natbib}
\usepackage{fp}

\definecolor{linkc}{rgb}{0, 0.44, 0.74}
\definecolor{eqc}{rgb}{1, 0, 0}
\definecolor{newcitecolor}{rgb}{0,0.6,0}
\definecolor{mygreen}{RGB}{34,139,34}
\definecolor{mylightblue}{RGB}{0,162,230}
\definecolor{deepyellow}{RGB}{255, 215, 0}
\definecolor{catgray}{gray}{0.92}
\definecolor{pearDark}{RGB}{171, 195, 87}
\definecolor{codebg}{RGB}{245, 245, 245}
\definecolor{keywordcolor}{RGB}{0, 0, 153}
\definecolor{commentcolor}{RGB}{34, 139, 34}
\definecolor{stringcolor}{RGB}{163, 21, 21}
\definecolor{numbercolor}{RGB}{128, 128, 128}

\def\onedot{\futurelet\@let@token\@onedot}
\def\@onedot{\ifx\@let@token.\else.\null\fi\xspace}

\makeatletter
\def\blfootnote#1{\xdef\@thefnmark{}\@footnotetext{\scriptsize #1}}
\makeatother

\hypersetup{
    colorlinks=true,
    urlcolor=shengshublue,
}

\let\cite\citep

\usepackage{amsmath,amsfonts,bm}

\def\eqref#1{equation~\ref{#1}}

\def\1{\bm{1}}

\def\vepsilon{{\bm{\epsilon}}}

\def\vx{{\bm{x}}}

\DeclareMathAlphabet{\mathsfit}{\encodingdefault}{\sfdefault}{m}{sl}
\SetMathAlphabet{\mathsfit}{bold}{\encodingdefault}{\sfdefault}{bx}{n}

\def\gS{{\mathcal{S}}}

\newcommand{\E}{\mathbb{E}}

\newcommand{\KL}{D_{\mathrm{KL}}}

\title{minWM: A Full-Stack Open-Source Framework for Real-Time Interactive Video World Models}

\author{%
  \textbf{Min Zhao}$^{1,2~\P*}$, 
    \textbf{Hongzhou Zhu}$^{1,2~*}$,
    \textbf{Bokai Yan}$^{1,3·*}$, \textbf{Zihan Zhou}$^{1,3~*}$, 
    \textbf{Yimin Chen}$^{1~\ddagger*}$,
    \textbf{Honglie Wang}$^{1~*}$, \\
    \textbf{Wenqiang Sun}$^{4~*}$, 
    \textbf{Zhengwei Fang}$^{2~}$,
    \textbf{Zizheng Xun}$^{2~}$,
    \textbf{Zihao Li}$^{2~}$, 
    \textbf{Kaiwen Zheng}$^{1,2}$,
    \textbf{Guande He}$^{5}$,
    \textbf{Xiao Yang}$^{2}$, \\ 
  \textbf{Chongxuan Li}$^{3}$, \textbf{Fan Bao}$^{1~\dagger}$, \textbf{Jun Zhu}$^{1,2~\dagger}$\\ 
  $^1$ShengShu\,
  $^2$THU\,
  $^3$RUC\,
  $^4$HKUST\,
  $^5$UT-Austin\\
  {\small $^*$ Equal contribution. \quad $\P$ Project lead. \quad $\ddagger$ Infra lead. \quad $^\dagger$ Advisor.}\\
\texttt{gracezhao1997@gmail.com;}
\texttt{dcszj@tsinghua.edu.cn}
}

\begin{abstract}
Recent video diffusion foundation models have achieved remarkable progress in high-quality video generation, yet turning them into real-time interactive video world models remains challenging. Interactive world models require controllable, causal, and low-latency rollout, which in practice demands a full pipeline spanning data construction, controllable fine-tuning, autoregressive training, few-step distillation, and streaming inference. In this work, we present \textbf{minWM}, a full-stack open-source framework for building real-time interactive video world models. minWM provides an end-to-end pipeline that converts existing bidirectional T2V/TI2V video foundation models into camera-controllable few-step autoregressive world models. Specifically, minWM first fine-tunes a bidirectional video diffusion model with camera control, and then applies the Causal Forcing / Causal Forcing++ pipeline, including AR diffusion training, causal ODE or causal consistency distillation, and asymmetric DMD, to distill it into a few-step autoregressive generator for low-latency rollout. The framework is modular and architecture-extensible: we instantiate it on representative open backbones, including Wan2.1-T2V-1.3B and HY1.5-TI2V-8B, covering both cross-attention-based condition injection and MMDiT-style architectures. minWM also supports adapting existing video world models, such as HY-WorldPlay, to new data distributions, training recipes, and latency targets. Beyond releasing runnable scripts, checkpoints, documentation, and inference code, we provide practical ablations on camera trajectory quality, controllability training steps, and minimal batch-size requirements. We hope minWM serves as a reproducible and extensible recipe for building and adapting real-time interactive video world models.
Project Page: \textbf{\href{https://github.com/shengshu-ai/minWM}{\textcolor{shengshublue}{https://github.com/shengshu-ai/minWM}}}.
\end{abstract}

\begin{document}
\maketitle

\section{Introduction}

Recent advances in diffusion-based video generation have produced powerful text-to-video (T2V) and text-and-image-to-video (TI2V) foundation models capable of synthesizing high-quality and temporally coherent videos~\cite{videoworldsimulators2024,bao2024vidu,yang2024cogvideox,lin2024open,zheng2024open,wan2025wan,kong2024hunyuanvideo}. These models provide strong generative priors for visual appearance, motion, and scene evolution, and therefore offer a promising starting point for building video world models. However, a high-quality offline video generator is not yet an interactive world model. An interactive video world model should support causal rollout, respond to user actions such as camera trajectories, and generate future frames with sufficiently low latency for real-time interaction~\cite{sun2025worldplay,genie3,tang2025hunyuan,mao2025yume,feng2025vidarc,huang2025live,sun2025streamavatar,hong2025relic,ye2025yan,xiang2025pan,he2025matrix,shin2025motionstream}.

Although recent works have explored autoregressive (AR) diffusion distillation to convert existing video foundation models into real-time interactive world models~\cite{yin2025slow,lin2025diffusion,huang2025self,zhu2026causal,zhao2026causal,yang2025towards}, these techniques remain scattered across separate pipelines. As a result, building an interactive video world model still requires substantial effort in data construction, controllable fine-tuning, AR training, few-step distillation, post-training alignment, and inference. \emph{A unified, reproducible, and extensible framework for this full pipeline is still missing}.

\begin{figure}[htbp]
    \centering
    \includegraphics[width=\textwidth]{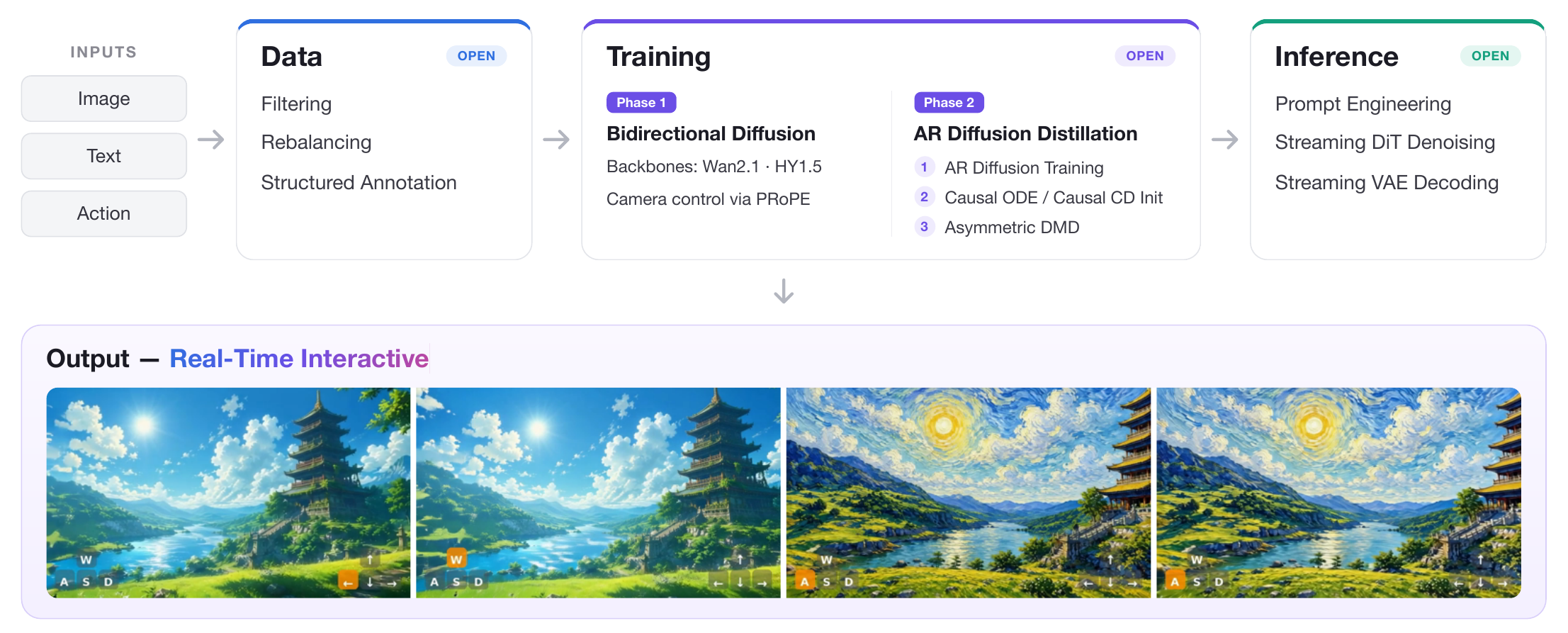}
    \caption{\textbf{Overview of minWM.}
minWM is a full-stack pipeline that converts T2V/TI2V foundation models into camera-controllable few-step autoregressive world models, covering data construction, controllable fine-tuning, AR training, distillation, and low-latency inference.}
    
    \label{fig:pipeline}
\end{figure}

To this end, we present \textbf{minWM}, a full-stack open-source framework for building real-time interactive video world models. Instead of releasing a single trained checkpoint, minWM provides a reproducible end-to-end pipeline that converts existing T2V or TI2V video foundation models into camera-controllable few-step autoregressive video world models. The framework covers the complete workflow, including data construction, camera-controllable fine-tuning, autoregressive diffusion training, few-step distillation, and low-latency inference. Its modular design allows researchers to plug in different video backbones, control signals, training recipes, and inference configurations, making minWM easy to reproduce, adapt, and extend.

Concretely, minWM follows a two-phase recipe. First, it fine-tunes a bidirectional video diffusion backbone on camera-annotated or camera-generated video data, enabling the model to follow prescribed camera trajectories while preserving the visual quality of the original foundation model~\cite{li2026cameras}. Second, it applies Causal Forcing~\cite{zhu2026causal} or Causal Forcing++~\cite{zhao2026causal} to transform the camera-controllable multi-step bidirectional model into a few-step autoregressive generator. This stage consists of teacher-forcing AR diffusion training~\cite{teng2025magi}, causal ODE~\cite{zhu2026causal} or causal consistency distillation~\cite{zhao2026causal} initialization, and asymmetric DMD~\cite{wang2023prolificdreamer,luo2023diff,yin2024one,yin2025slow} post-training with self-rollout~\cite{huang2025self}. The resulting model supports camera-controllable autoregressive video generation with few-step inference, making it suitable for low-latency interactive applications.

We instantiate minWM on representative open video backbones, including Wan2.1-T2V-1.3B~\cite{wan2025wan} and HY1.5-TI2V-8B~\cite{kong2024hunyuanvideo}. These instantiations demonstrate two practical usages of the framework. First, minWM provides a complete conversion pipeline that starts from a bidirectional T2V or TI2V foundation model and progressively turns it into a real-time, camera-controllable autoregressive video world model. By releasing intermediate checkpoints for each training stage, minWM allows researchers to resume, modify, or extend the pipeline from any stage. Second, minWM supports adapting existing video world models, such as HY-WorldPlay~\cite{sun2025worldplay}, to new data distributions, training recipes, or latency targets through fine-tuning and distillation. Beyond final generation results, we further report practical ablations on camera trajectory quality of the dataset, controllability training steps, and minimal batch-size requirements, providing actionable guidance for reproducible interactive world-model training. The overall pipeline is illustrated in Fig. \ref{fig:pipeline}.

Our contributions are summarized as follows:

\begin{itemize}
    \item We release \textbf{minWM}, a fully open-source end-to-end pipeline for building real-time interactive video world models. The pipeline covers camera-conditioned data construction, controllable fine-tuning of bidirectional video diffusion models, and Causal Forcing / Causal Forcing++ distillation, including AR diffusion training, causal ODE or causal consistency distillation initialization, asymmetric DMD post-training, and low-latency inference. 

    \item We show that minWM is architecture-general and can convert multiple types of video foundation models into camera-controllable few-step autoregressive world models. We instantiate the framework on representative open backbones, including Wan2.1-T2V-1.3B with cross-attention-based condition injection and HY1.5-TI2V-8B with an MMDiT-style architecture~\cite{esser2024scaling}.

    \item We further support the adaptation of existing video world models, such as HY-WorldPlay, to new data distributions, training recipes, and latency targets. Together with practical ablations on camera trajectory quality, controllability training steps, and minimal batch-size requirements, minWM provides a reproducible and extensible recipe for building and adapting interactive video world models.
\end{itemize}
\section{Method}
\label{sec: method}
In this section, we present how to convert a text-to-video (T2V) or text-and-image-to-video (TI2V) multi-step bidirectional diffusion model into a camera-controllable few-step autoregressive (AR) video generator.The pipeline consists of two major phases: first, \emph{Camera Control Training for Bidirectional Diffusion Models} (Sec.~\ref{sec: method-bidirectional}), which equips the multi-step bidirectional diffusion model with camera controllability; and second, \emph{AR Diffusion Distillation for Real-Time Interactive Models} (Sec.~\ref{sec: method-ar}) via Causal Forcing~\cite{zhu2026causal} or Causal Forcing++~\cite{zhao2026causal}, which transforms the model into a real-time interactive AR model.

\subsection{Camera-Controllable Training for Bidirectional Diffusion Models}
\label{sec: method-bidirectional}

In this section, we fine-tune the T2V or TI2V bidirectional diffusion model into a camera-controllable bidirectional diffusion model. We adopt PRoPE~\cite{li2026cameras} as the injection method for camera parameters.

Specifically, given a video clip with camera parameters 
$\{(K_i,T_i^{cw})\}_{i=1}^{N}$, where $K_i$ denotes the intrinsic matrix and 
$T_i^{cw}\in SE(3)$ denotes the world-to-camera extrinsic transformation of frame $i$, PRoPE represents each camera by its lifted projective matrix
\[
\widetilde P_i=
\begin{bmatrix}
[K_i\;0]T_i^{cw}\\
e_4^\top
\end{bmatrix}\in\mathbb{R}^{4\times 4},
\qquad
e_4=(0,0,0,1)^\top .
\]
For a token $t$ belonging to frame $i(t)$ with spatial coordinate $(x_t,y_t)$, PRoPE constructs a block-diagonal transformation
\[
D_t^{\mathrm{PRoPE}}
=
\begin{bmatrix}
I_{d/8}\otimes \widetilde P_{i(t)} & 0\\
0 &
\begin{bmatrix}
\mathrm{RoPE}_{d/4}(x_t)&0\\
0&\mathrm{RoPE}_{d/4}(y_t)
\end{bmatrix}
\end{bmatrix}.
\]
This transformation is injected into self-attention in the GTA form:
\[
\mathrm{Attn}_{\mathrm{PRoPE}}(Q,K,V)
=
D^{\mathrm{PRoPE}}\odot
\mathrm{Attn}\!\left(
(D^{\mathrm{PRoPE}})^\top\odot Q,\,
(D^{\mathrm{PRoPE}})^{-1}\odot K,\,
(D^{\mathrm{PRoPE}})^{-1}\odot V
\right).
\]
Consequently, the attention interaction between tokens $t_1$ and $t_2$ explicitly depends on the relative projective transformation
\[
\widetilde P_{i(t_1)}\widetilde P_{i(t_2)}^{-1}
=
\begin{bmatrix}
K_{i(t_1)}&0\\
0&1
\end{bmatrix}
T_{i(t_1)}^{cw}
\bigl(T_{i(t_2)}^{cw}\bigr)^{-1}
\begin{bmatrix}
K_{i(t_2)}^{-1}&0\\
0&1
\end{bmatrix},
\]
thereby jointly encoding relative camera intrinsics and camera poses. This allows the bidirectional diffusion backbone to condition on camera trajectories while preserving the original self-attention generative structure.

\subsection{AR Diffusion Distillation for Real-Time Interactive Video World Models}
\label{sec: method-ar}

In this section, we can either adopt Causal Forcing~\cite{zhu2026causal} or Causal Forcing++~\cite{zhao2026causal} to transform the camera-controllable multi-step bidirectional diffusion model obtained in Sec.~\ref{sec: method-bidirectional} into a camera-controllable few-step AR model. This distillation pipeline consists of three stages: \emph{\textbf{(1) Stage 1: AR diffusion training; (2) Stage 2: causal ODE initialization or causal CD initialization; and (3) Stage 3: asymmetric DMD.}} 

\paragraph{Stage 1: AR diffusion training.} Starting from a multi-step bidirectional diffusion model, Causal Forcing~\cite{zhu2026causal} first fine-tunes it into an AR diffusion model via teacher forcing~\cite{teng2025magi}. This is achieved by concatenating the clean video with its noisy counterpart and training the model under a causal attention mask. The resulting model already possesses autoregressive generation capability, but still suffers from two limitations: (1) it requires multi-step generation, leading to high latency; and (2) due to exposure bias induced by autoregression~\cite{yin2025slow}, its quality remains inferior to that of bidirectional diffusion models. These limitations motivate the subsequent distillation strategy.

\paragraph{Stage 2 (option a): causal ODE initialization.} Causal Forcing~\cite{zhu2026causal} points out that using an AR diffusion model to supervise an AR few-step model, as the subsequent DMD initialization, helps improve generation quality. This AR diffusion model generates a large number of intermediate denoising trajectories, namely PF-ODE trajectories~\cite{song2020score}. Then, over a predefined few-step timestep set $\gS$, a timestep $t$ is randomly sampled, and the few-step model $G_\theta$ is trained by regressing from the noisy intermediate frame $\vx_t^i$ to the clean frame $\vx_0^i$:
\begin{align}
    \theta^*=\arg\min_\theta \mathbb{E}_{\vx_{\mathrm{gt}}^{<i},\, t,\, i,\, \vx_t^i}\left[\,\|G_\theta(\vx_t^i,\vx_{\mathrm{gt}}^{<i}, t)-\vx_0^i\|^2\,\right],
\end{align}
where $\vx_{\mathrm{gt}}^{<i}$ denotes the historical prefix formed by real data. The model trained in this way can already perform few-step autoregressive generation, but its quality is constrained by the AR diffusion model and remains inferior to that of the bidirectional model, thus motivating the need for asymmetric DMD (i.e., Stage 3).

\paragraph{Stage 2 (option b): causal CD initialization.} ODE distillation requires generating offline ODE data, which is both time-consuming and storage-intensive. To eliminate this data curation time and the storage overhead of ODE trajectories, Causal Forcing++~\cite{zhao2026causal} further replaces this stage with the theoretically equivalent causal consistency distillation~\cite{song2023consistency}, namely causal CD:
\begin{align}
\label{eq:causal-cd}
\theta^*=\arg\min_\theta \mathbb{E}_{\vx_\text{gt},\,\vepsilon,\,t,\,i}\Big[w(t)\,d\big(G_\theta(\vx_t^i,\vx_\text{gt}^{<i}, t),\; G_{\theta^-}(\hat{\vx}^i_{t-\Delta t},\vx_\text{gt}^{<i}, t-\Delta t)\big)\Big],
\end{align}
where $\hat{\vx}^i_{t-\Delta t}$ is obtained by a single ODE step from $\vx_t^i$ using the AR teacher conditioned on $\vx_\text{gt}^{<i}$, $\theta^-$ is the EMA of $\theta$ with stop-gradient, $w(\cdot)$ is a timestep-dependent weight, and $d(\cdot,\cdot)$ is a distance under a pre-defined norm. A model trained in this way is equivalent to one obtained via causal ODE distillation~\cite{zhu2026causal}.

\paragraph{Stage 3: asymmetric DMD.} The resulting few-step AR model is already capable of real-time generation, but since the AR teacher has limited generation quality, it inherits this limitation. Therefore, a final asymmetric DMD stage is applied using the bidirectional diffusion model, aligning the few-step AR model with the high-quality distribution of the bidirectional teacher~\cite{yin2025slow,huang2025self}: the student model is initialized from the above few-step AR model, self-rolls out to generate a full video sequence $\tilde{\vx}$, and is then optimized with the standard DMD gradient as follows~\cite{wang2023prolificdreamer,yin2024one}:
\begin{align}
\label{eq:dmd}
\nabla_\theta\E_t[\KL(p_{\theta,t}(\tilde{\vx}_t)||p_{\text{data},t}(\tilde{\vx}_t))]
   = -\E_{\tilde{\vx}, t, \tilde{\vx}_t}[(s_\text{real}(\tilde{\vx}_t,t)-s_\text{fake}(\tilde{\vx}_t,t))\frac{\partial \tilde{\vx}}{\partial \theta}].
\end{align}
Here, $\tilde{\vx}$ is perturbed into $\tilde{\vx}_t\sim  p_\theta(\tilde{\vx})$ through the forward diffusion process, thereby inducing the marginal distribution $p_{\theta,t}(\tilde{\vx}_t)$. The score of $\tilde{\vx}_t$ in the data distribution is estimated by a frozen diffusion model $s_{\text{real}}$, whereas the score of $\tilde{\vx}_t$ in $p_\theta(\tilde{\vx})$ is estimated by an online-trained diffusion model $s_{\text{fake}}$.

\paragraph{Camera-controllable distillation.} For camera-controllable video world models, we only need to instantiate the Causal Forcing series from the camera-controllable multi-step bidirectional diffusion model. Specifically, in Stage 1, the AR diffusion model is initialized from the camera-controllable multi-step bidirectional diffusion model obtained in Sec.~\ref{sec: method-bidirectional} and is still trained on camera-controllable data. In Stage 2, when collecting causal ODE data, the AR diffusion model also takes the camera condition as input to solve the PF-ODE; similarly, causal CD is trained on camera-controllable data. In Stage 3, the student model takes not only the text condition but also the camera condition for self-rollout, and the same camera condition is also fed into $s_\text{real}$ and $s_\text{fake}$, which are initialized from the camera-controllable multi-step bidirectional diffusion model obtained in Sec.~\ref{sec: method-bidirectional}. In summary, all involved models are camera-controllable.
\section{Experiments}
\label{sec:exp}

In this section, we present the detailed experimental setup, generation results, and ablation studies on key training factors.

\subsection{Setup}

We train two models, Wan2.1-T2V-1.3B~\cite{wan2025wan} and HY1.5-TI2V-8B~\cite{kong2024hunyuanvideo}, to generate videos of resolution $480\times 832$ with 77 frames. The autoregressive chunk size is set to 4 latent frames. For few-step distillation, we use 4 steps following Causal Forcing~\cite{zhu2026causal}. Roughly speaking, unless otherwise specified, for the HY1.5-based training, we use a batch size of 32 and a learning rate of $1\times 10^{-5}$; the bidirectional model is trained for 8K steps, followed\footnote{This 8K-step model is used as $s_{\text{real}}$ and $s_{\text{fake}}$ in Causal Forcing Stage 3, whereas Stage 1 is initialized from the 5K-step model.} by 4K steps for Causal Forcing Stage 1, 1.5K steps for Stage 2, and 500 steps for Stage 3. For the Wan2.1-based training, we use a batch size of 32 and a learning rate of $2\times 10^{-6}$; the bidirectional model is trained for 5K steps, followed by 4K steps for Causal Forcing Stage 1, 2K steps for Stage 2, and 200 steps for Stage 3. For details on the data, please refer to Sec.~\ref{sec:ablation}.

\begin{table}[t]
    \centering
    \caption{\textbf{First-frame latency of different HY1.5 and Wan2.1 models.} We report the first-frame latency on a single A800 GPU. VAE-related time is excluded.}
    \label{tab:latency}
    \begin{tabular}{lccc}
        \toprule
        Base model & Model type & First-frame latency (s) & Speedup over multi-step bidirectional \\
        \midrule
        HY1.5~\cite{kong2024hunyuanvideo} & Multi-step bidirectional & 771.041 & $1.00\times$ \\
        HY1.5 & Multi-step AR & 81.014 & $9.52\times$ \\
        HY1.5 & Few-step AR & 3.446 & $223.75\times$ \\
        \midrule
        Wan2.1 & Multi-step bidirectional & 269.055 & $1.00\times$ \\
        Wan2.1~\cite{wan2025wan} & Multi-step AR & 28.651 & $9.39\times$ \\
        Wan2.1 & Few-step AR & 1.137 & $236.64\times$ \\
        \bottomrule
    \end{tabular}
\end{table}

\subsection{Results}
In this section, we present the final results of applying the minWM framework to Wan2.1 and HY1.5. We first report the first-frame latency on the single A800 GPU excluding the VAE-related time, and then show several generated video samples.

\paragraph{Few-step AR models substantially reduce the first-frame latency.
} As shown in Tab.~\ref{tab:latency}, minWM substantially reduces the first-frame latency of both base models. In particular, the final few-step AR model achieves a $223.75\times$ first-frame latency reduction over the multi-step bidirectional HY1.5 baseline, and a $236.64\times$ first-frame latency reduction over the multi-step bidirectional Wan2.1 baseline. Notably, since the bidirectional model generates the entire sequence at once, its first-frame latency is naturally much higher than that of the AR model, which generates the first frame first and then continues to generate subsequent frames. In practical deployment scenarios, the low first-frame latency of the AR model allows users to start watching while generation is still ongoing, thereby reducing perceived waiting time.

\begin{figure}[htbp]
    \centering
    \includegraphics[width=\textwidth]{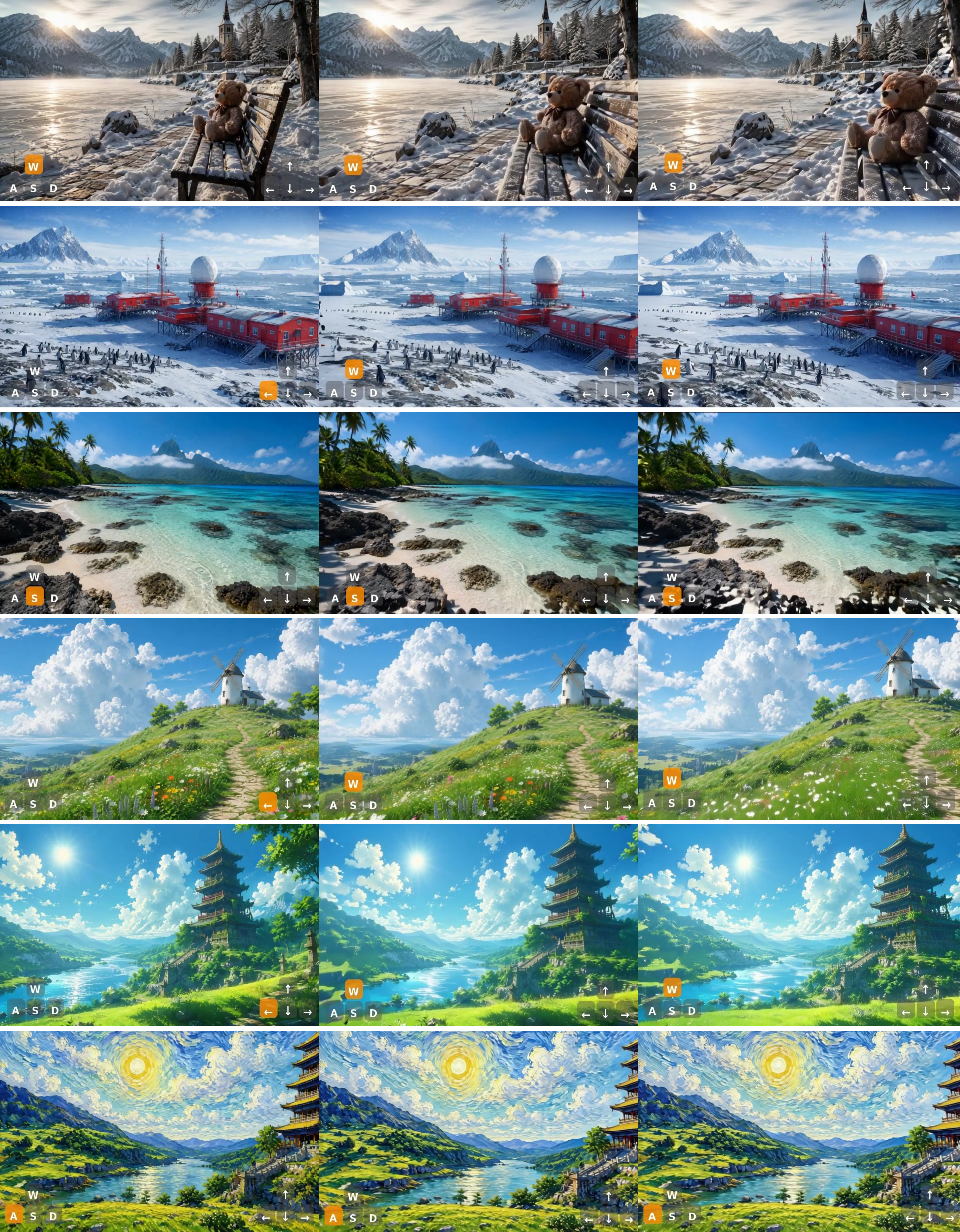}
    \caption{\textbf{Camera-controllable generation with the distilled few-step AR model.}
    The model supports generation under different camera actions, showing that the distillation algorithm effectively preserves the camera controllability of the base model.}
    
    \label{fig:main}
\end{figure}

\paragraph{Few-step AR models preserve camera-controllable generation capability.} As shown in Fig.~\ref{fig:main}, the model is capable of camera-controllable generation and supports changing the camera action, demonstrating the effectiveness of the distillation algorithm in preserving the model's controllability.

\subsection{Ablation Studies}
\label{sec:ablation}
In this section, we examine key factors encountered during training and present the corresponding ablation studies.

\paragraph{Training data.}

\begin{figure}
  \centering

  \begin{subfigure}{\linewidth}
    \includegraphics[width=\linewidth]{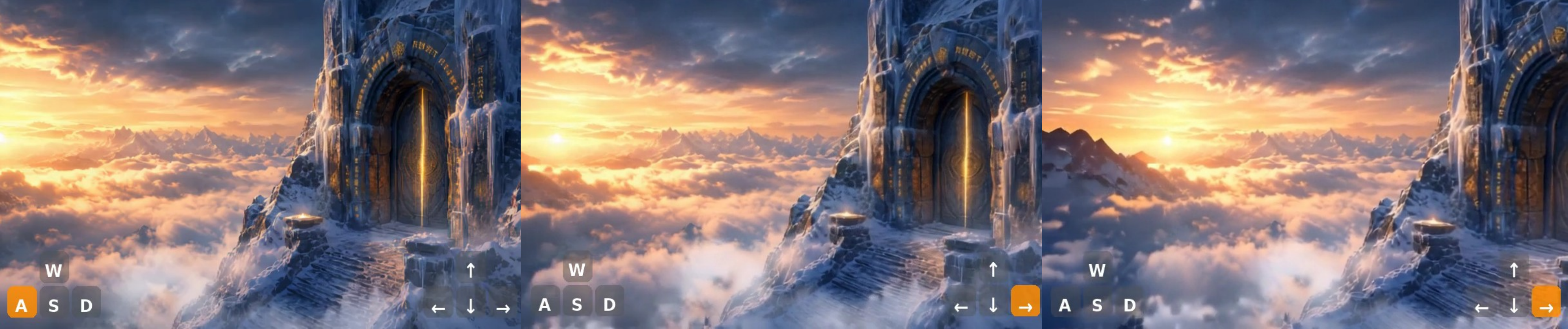}
    \caption{\textbf{Training with estimated camera poses.}
In our experiments, models trained directly on SpatialVid~\cite{wang2025spatialvid} data did not achieve reliable camera-controllable generation under our current setup, even after additional data filtering. We hypothesize that this may be related to the use of perception-estimated camera poses, which motivates our exploration of datasets with effectively ground-truth trajectories.}
  \end{subfigure}

  \vspace{0.5em}

  \begin{subfigure}{\linewidth}
    \includegraphics[width=\linewidth]{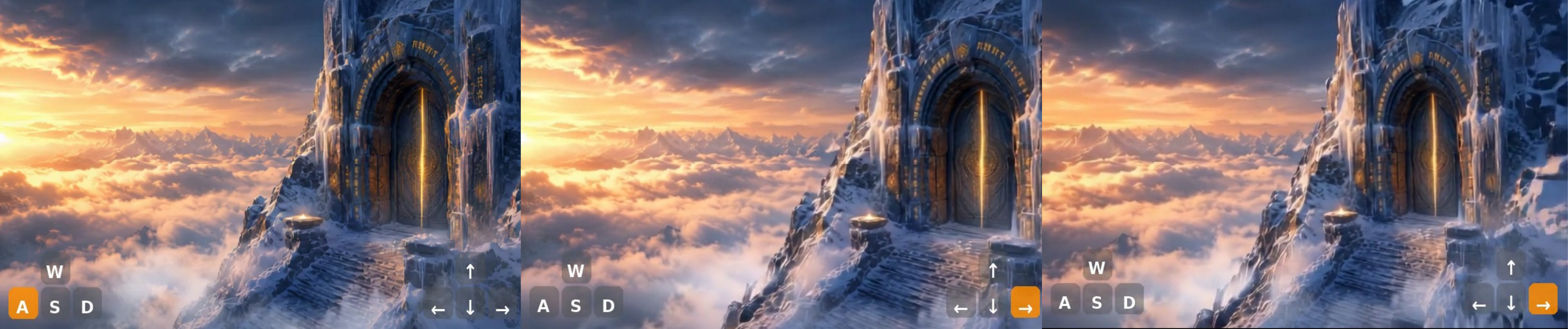}
    \caption{\textbf{Training with reconstructed scenes and rendered trajectories.}
    By reconstructing scenes from DL3DV~\cite{ling2024dl3dv} and rendering videos along prescribed camera trajectories, the model successfully learns camera-controllable generation, indicating the importance of accurate camera trajectories.}
  \end{subfigure}

  \vspace{0.5em}

  \begin{subfigure}{\linewidth}
    \includegraphics[width=\linewidth]{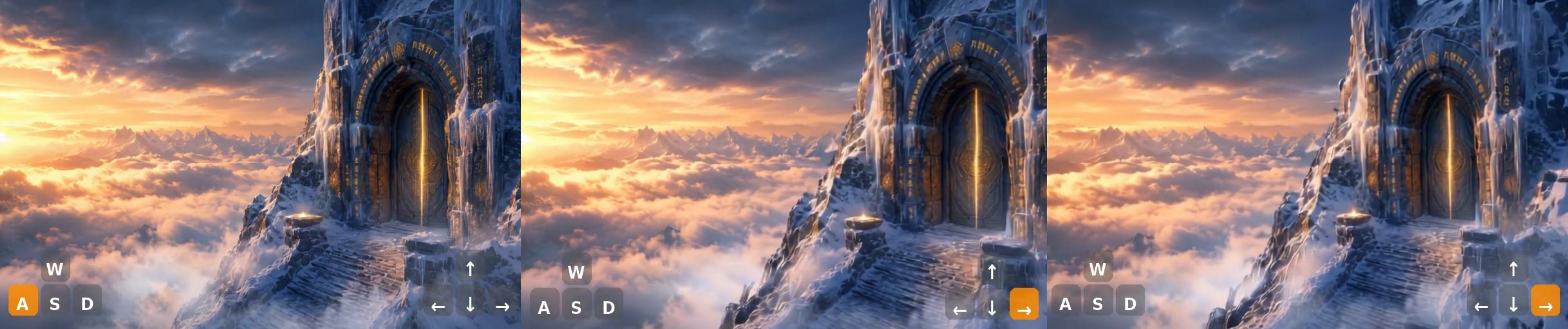}
    \caption{\textbf{Training with WorldPlay-generated trajectories.}
    For the open-source setting, we construct videos from OpenVid~\cite{nan2024openvid} and other image sources using WorldPlay~\cite{sun2025worldplay} with specified camera trajectories, which likewise enables the model to learn camera controllability.}
  \end{subfigure}

  \caption{\textbf{Effect of training data on camera-controllable generation.}
Under our current setup, directly training with SpatialVid did not yet yield reliable camera-controllable generation. We therefore construct datasets with effectively ground-truth camera trajectories, either through 3D reconstruction and re-rendering or WorldPlay-based generation, which enable the model to learn camera controllability.}
\label{fig:ablation-data}
\end{figure}

We first attempted to train on SpatialVid~\cite{wang2025spatialvid} data. Under our current training setup, however, models trained in this way, including both HY1.5~\cite{kong2024hunyuanvideo} and Wan2.1~\cite{wan2025wan}, did not yet achieve reliable camera-controllable generation, as illustrated in Fig.~\ref{fig:ablation-data}(a). Even with additional data filtering, the model still struggled to perform accurate camera control in our experiments. We hypothesize that this may be related to the use of perception-estimated camera poses, which can introduce pose noise or trajectory inconsistency compared with ground-truth trajectories. \textit{This result should be interpreted as a limitation of our current SpatialVid-based training attempt rather than a conclusion that SpatialVid is unsuitable for this task.} We leave improved filtering, pose refinement, and more systematic SpatialVid-based training to future work.

Based on this observation, we argue that ground-truth camera poses are crucial. We therefore adopt a 3D reconstruction and re-rendering strategy: we reconstruct scenes from the DL3DV~\cite{ling2024dl3dv} dataset and then render videos along prescribed camera trajectories. With this data, the model successfully learns camera controllability, as illustrated in Fig.~\ref{fig:ablation-data}(b).

For the open-source version, we adopt another dataset construction strategy: we sample images from OpenVid~\cite{nan2024openvid} and other sources, and use WorldPlay~\cite{sun2025worldplay} to generate videos following specified camera trajectories. This also provides effective ground-truth trajectories, and the model can likewise learn camera controllability, as illustrated in Fig.~\ref{fig:ablation-data}(c).

\paragraph{Training steps.} 

\begin{figure}
  \centering

  \begin{subfigure}{\linewidth}
    \includegraphics[width=\linewidth]{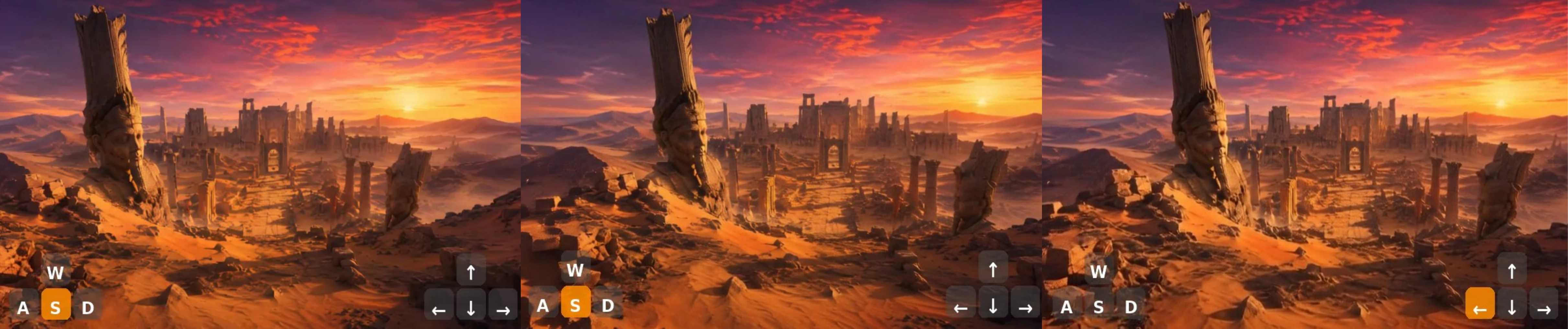}
    \caption{\textbf{Early-stage training.}
    After only one to two thousand training steps, the HY1.5-based bidirectional model has not yet acquired effective camera controllability.}
  \end{subfigure}

  \vspace{0.5em}

  \begin{subfigure}{\linewidth}
    \includegraphics[width=\linewidth]{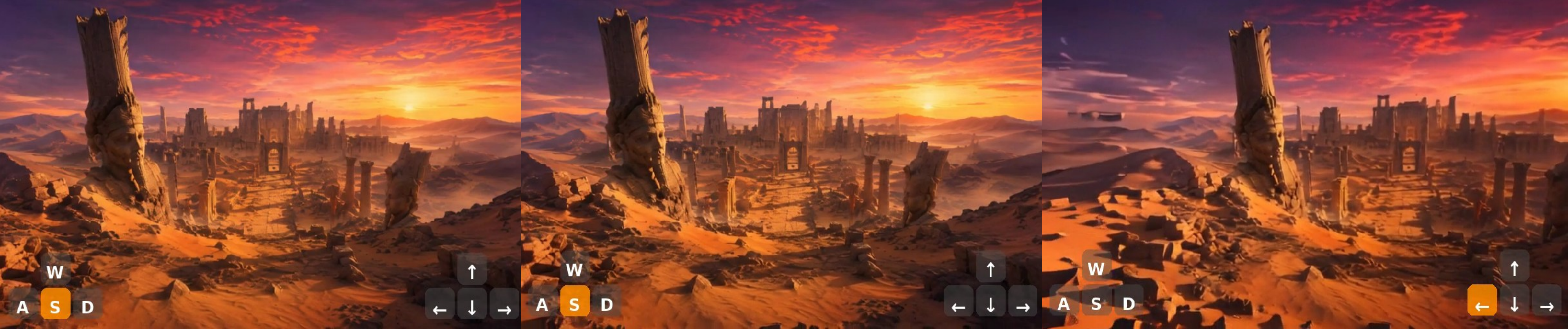}
    \caption{\textbf{Emerging controllability.}
    After around five thousand training steps, the model begins to respond to camera-control signals, but the controllability is still unstable.}
  \end{subfigure}

  \vspace{0.5em}

  \begin{subfigure}{\linewidth}
    \includegraphics[width=\linewidth]{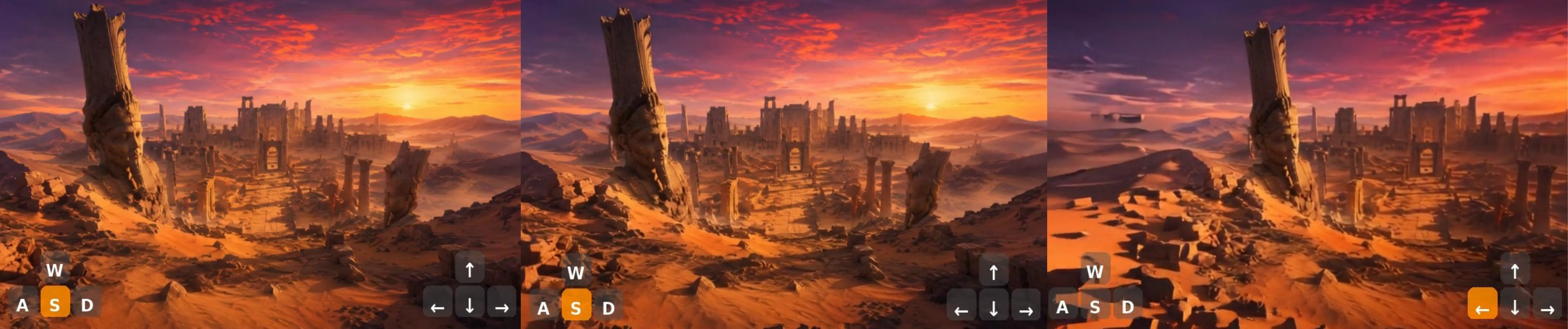}
    \caption{\textbf{Strong controllability.}
    After eight thousand training steps, the model achieves substantially stronger and more reliable camera-controllable generation.}
  \end{subfigure}

  \caption{\textbf{Effect of training steps on camera-controllable generation.}
  Using HY1.5 as an example, we observe that camera controllability emerges progressively during training: the model is largely uncontrollable at one to two thousand steps, starts to acquire controllability around five thousand steps, and reaches strong controllability after eight thousand steps.}
  \label{fig:ablation-steps}

\end{figure}

Taking HY1.5 as an example, we further report the number of training steps required for the bidirectional diffusion model to acquire camera controllability. We find that after only one to two thousand training steps, the model remains completely uncontrollable, as illustrated in Fig.~\ref{fig:ablation-steps}(a). After around five thousand steps, the model begins to exhibit camera controllability, as illustrated in Fig.~\ref{fig:ablation-steps}(b). After eight thousand steps, the model achieves strong controllability, as illustrated in Fig.~\ref{fig:ablation-steps}(c).

\paragraph{Minimal batch size.}

\begin{figure}
  \centering

  \begin{subfigure}{\linewidth}
    \includegraphics[width=\linewidth]{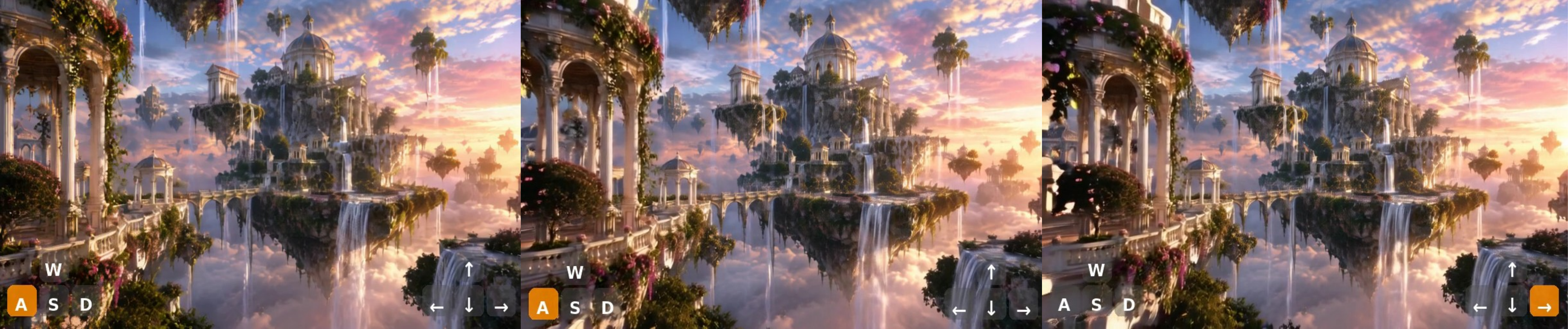}
    \caption{\textbf{Training with very small batch sizes.}
    When the batch size is smaller than 4, the Wan2.1-based model often fails to learn camera-controllable generation.}
  \end{subfigure}

  \vspace{0.5em}

  \begin{subfigure}{\linewidth}
    \includegraphics[width=\linewidth]{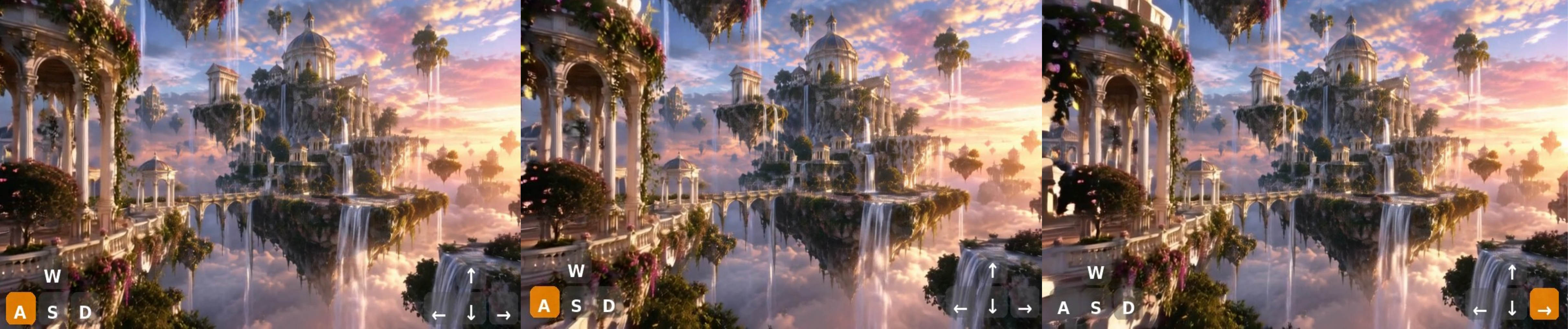}
    \caption{\textbf{Training with batch size 8.}
    With a batch size of 8, the model's controllability improves substantially, but remains somewhat unstable.}
  \end{subfigure}

  \vspace{0.5em}

  \begin{subfigure}{\linewidth}
    \includegraphics[width=\linewidth]{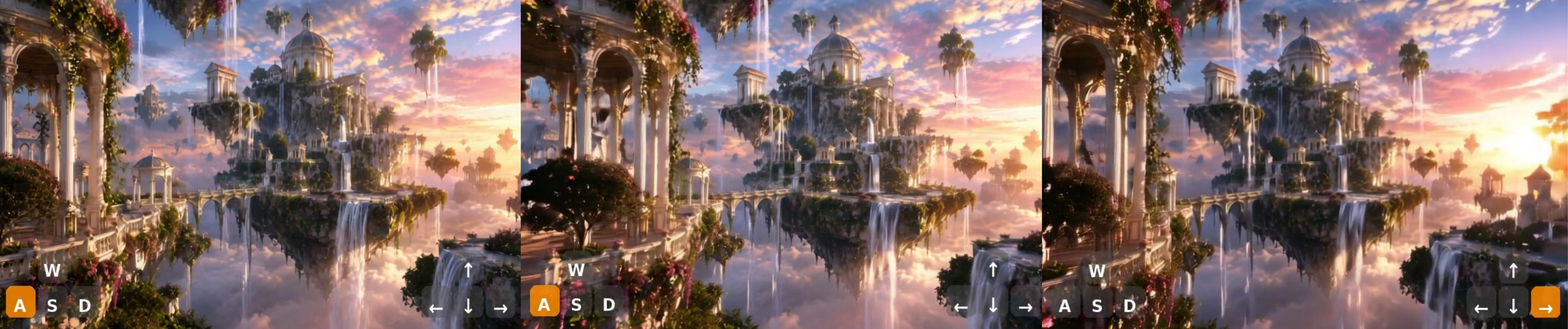}
    \caption{\textbf{Training with batch size 16.}
    With a batch size of 16, the full training pipeline can be successfully completed with high controllability.}
  \end{subfigure}

  \caption{\textbf{Effect of batch size on camera-controllable generation.}
  Using Wan2.1 as an example, we find that batch size critically affects camera-control training: batch sizes below 4 often fail to learn controllability, batch size 8 substantially improves controllability but remains unstable, while batch size 16 enables successful training with high controllability.}
  \label{fig:ablation-bs}

\end{figure}

Taking Wan2.1 as an example, we investigate the minimum batch size required for learning camera controllability, aiming to facilitate research under limited computational budgets. We find that when the batch size is smaller than 4, the model often fails to learn camera controllability, as illustrated in Fig.~\ref{fig:ablation-bs}(a). With a batch size of 8, the model's controllability improves substantially, but remains somewhat unstable, as illustrated in Fig.~\ref{fig:ablation-bs}(b). With a batch size of 16, the full training pipeline can be successfully completed with high controllability, as illustrated in Fig.~\ref{fig:ablation-bs}(c).

\section{Conclusion and the Future Work}
We propose minWM, a full-stack open-source framework for video world models. It supports fine-tuning bidirectional T2V or TI2V models for camera-controllable generation, as well as distilling them into real-time interactive AR models. minWM currently supports HY1.5~\cite{kong2024hunyuanvideo} and Wan2.1~\cite{wan2025wan}. In the future, we plan to support additional control conditions beyond camera control, such as pose, and to extend the framework to more models.

\clearpage

{
  \small
  \bibliographystyle{unsrt}
  \bibliography{main}
}

\end{document}